\documentclass[letterpaper, 10 pt, conference]{ieeeconf}
\IEEEoverridecommandlockouts                             
\usepackage{amsmath}
\usepackage{mathtools}
\usepackage{xcolor}
\usepackage{graphicx}
\usepackage{tikz}
\usepackage{textcomp}
\usepackage{hyperref}
\usepackage[hyphenbreaks]{breakurl}

\title{\LARGE \bf
Evaluation of Local Planner-Based Stanley Control in Autonomous RC Car Racing Series
}

\author{Máté Fazekas, Zalán Demeter, János Tóth, Ármin Bogár-Németh, Gergely Bári%
\thanks{The research was supported by the National Research, Development and Innovation Office. (2020-2.1.1-ED-2021-00162)}%
\thanks{All Authors with Széchenyi István University, H-9026 Győr, Hungary (e-mail: [mate.fazekas,zalan.demeter,janos.toth,armin.bogar-nemeth,gergely.bari]@humda.hu])}%
}

\newcommand\copyrighttext{%
  \footnotesize \textcopyright 2024 IEEE. Personal use of this material is permitted. Permission from IEEE must be obtained for all other uses, in any current or future media, including reprinting/republishing this material for advertising or promotional purposes, creating new collective works, for resale or redistribution to servers or lists, or reuse of any copyrighted component of this work in other works.
  DOI: \href{https://doi.org/10.1109/IV55156.2024.10588629}{10.1109/IV55156.2024.10588629}}
\newcommand\copyrightnotice{%
\begin{tikzpicture}[remember picture,overlay]
\node[anchor=south,yshift=10pt] at (current page.south) {\fbox{\parbox{\dimexpr\textwidth-\fboxsep-\fboxrule\relax}{\copyrighttext}}};
\end{tikzpicture}%
}

\begin{document}

\maketitle
\copyrightnotice
\thispagestyle{empty}
\pagestyle{empty}

\begin{abstract}

This paper proposes a control technique for autonomous RC car racing. The presented method does not require any map-building phase beforehand since it operates only local path planning on the actual LiDAR point cloud. Racing control algorithms must have the capability to be optimized to the actual track layout for minimization of lap time. In the examined one, it is guaranteed with the improvement of the Stanley controller with additive control components to stabilize the movement in both low and high-speed ranges, and with the integration of an adaptive lookahead point to induce sharp and dynamic cornering for traveled distance reduction. The developed method is tested on a 1/10-sized RC car, and the tuning procedure from a base solution to the optimal setting in a real F1Tenth race is presented. Furthermore, the proposed method is evaluated with a comparison to a more simple reactive method, and in parallel to a more complex optimization-based technique that involves offline map building the global optimal trajectory calculation. The performance of the proposed method compared to the latter, referring to the lap time, is that the proposed one has only $8\%$ lower average speed. This demonstrates that with appropriate tuning, a local planning-based method can be comparable with a more complex optimization-based one. Thus, the performance gap is lower than $10\%$ from the state-of-the-art method. Moreover, the proposed technique has significantly higher similarity to real scenarios, therefore the results can be interesting in the context of automotive industry. 

\end{abstract}

\section{INTRODUCTION}

In recent years, the automotive industry has undergone a revolutionary transformation with the emergence of autonomous driving technologies. The goal for safer, more efficient, and convenient transportation has led to the development of vehicles capable of navigating complex environments without direct human intervention \cite{Milakis:2017}. As these autonomous systems become more dominant, understanding the complex dynamics between technology, safety, and performance becomes a key question.

One critical aspect of the field of autonomous driving is the handling of grip limit – the state when the vehicle moves with the available maximum tire forces before losing control \cite{pacejka2005tire}. This factor becomes particularly significant when evaluating the capabilities and limitations of autonomous vehicles, especially in car racing series where the paramount aim of teams is to reach minimum lap time, which is equal to controlling the racecar at the grip limit continuously \cite{Betz:2022}.

The architecture of an autonomous driving system in a racing series can be developed in a wide range of methods, from the classic see-think-act process \cite{thrun:2002}, where separate blocks with clearly defined specific tasks are connected \cite{demeter:2024}. The other extreme is a completely end-to-end approach with a learning-based technique, such as reinforcement learning \cite{Evans2023},\cite{rl2024}. Furthermore, any kind of combination of this paradigm for any task can exist, e.g., for perception and mapping \cite{Venkatesh:2018}, path planning \cite{Minh:2019}, or control \cite{Ghignone_2023}.

Although the top racing pilots are certainly closer to the end-to-end paradigm \cite{bari_vision_2024}, teams in autonomous series could develop block-oriented types of control algorithms that can reach $>200 \ km/h $ average speed on a track \cite{https://doi.org/10.1002/rob.22153}. These methods branch out in terms of whether or not to integrate a pre-calculated offline map into the architecture. There is no doubt that a method that operates with a given map optimizes the ideal line on it and utilizes it as a trajectory in the online tracking phase can achieve higher performance than another that plans the trajectory to follow only online, e.g., from the actual LiDAR measurements. However, the research outcomes of the last have a significantly higher impact in the context of the automotive industry since the examined case is significantly closer to the real-life scenarios. Thus, an interesting research to study is how relating the results obtained with a control algorithm using only online planning compares to the results reached with methods using a predefined map and globally optimal trajectory.

This paper presents a control method that utilizes only online local path planning to control a 1/10-sized RC racecar fully autonomously. The method uses a LiDAR sensor to measure the track walls in a racetrack and calculates the actual centerline. The control algorithm is a modified Stanley-controller, which is tuned to the actual track to achieve minimum lap time. The proposed method is tested with real RC cars, and the performance is evaluated with other control techniques in an F1Tenth competition.

The rest of the paper is organized as follows. The general architecture of autonomous systems and the examined control techniques are summarized in Section \ref{sec:overview_methods}. The developed method is presented in Section \ref{sec:our_methods}, while the details of the experimental vehicle are in Section \ref{sec:vehicle}, respectively. The tuning results and performance indicators of the proposed method and the comparison with other techniques can be found in Section \ref{sec:results}. Finally, the paper is concluded in Section \ref{sec:conclusion}.

\section{Overview of control approaches} \label{sec:overview_methods}

First, a brief overview of the architecture of an autonomous driving system and the control approaches are presented to locate the method we have developed and examined in the spectrum of possible implementations.

\subsection{Architecture of autonomous driving system}

\begin{figure}
    \centering
    \includegraphics[width=0.48\textwidth]{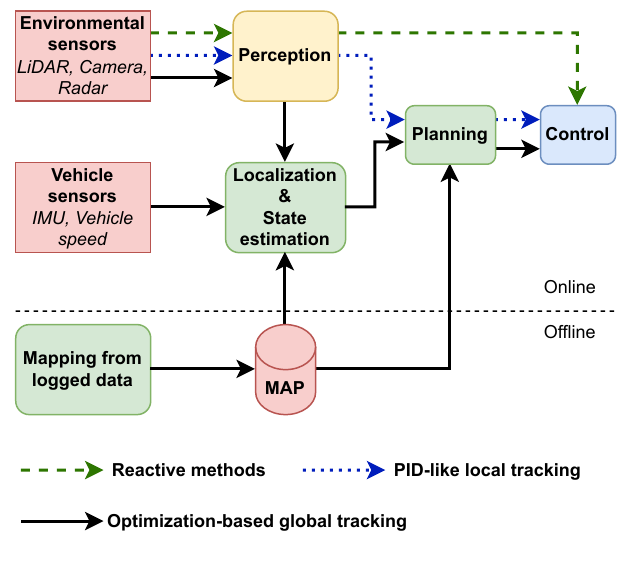}
    \caption{Block diagram of the autonomous system architecture and illustration of the connection of the examined control methods}
    \label{fig:AD_5_layers}
\end{figure}

The architecture of a fully autonomous driving system has 5 main consecutive layers. In the \textit{Perception}, the raw measurements of environmental sensors, e.g., LiDAR, camera, radar, are transformed into a high-level feature, such as lane or objects. These are used in the \textit{Mapping} layer to form regions such as track, border limits, and others. This map generation is often performed offline, while the \textit{Localization}, the process of estimating the vehicle's pose in the map frame, runs online. In this part of the architecture, the \textit{Estimation} of other quantities, such as velocity or yaw rate, is accomplished in parallel. The preferred trajectory to follow is determined in the \textit{Planning} layer, while at the end, the wheel angle and velocity \textit{Control} signals are calculated. The layers and their connection to form the various control methods are illustrated in Figure \ref{fig:AD_5_layers}.

\subsection{General control methods} \label{control_methods}

The vehicle can be controlled in several ways in connection with the implementation of the layers mentioned aforementioned. The options spread from methods for determining the steering angle directly from raw measurements to following a pre-calculated ideal trajectory. For the scope of our evaluation, the control methods are briefly arranged into three groups, from the simplest to the most complex.

\subsubsection{Reactive methods}

The techniques that neglect the \textit{Mapping-Estimation-Planning} layers and establish a unique and straightforward relation between the \textit{Perception} and \textit{Control} parts can be found in the first group. One of the most common of these methods is the Follow-The-Gap, which was originally developed for obstacle avoidance \cite{sezer:2012}, however, the pipes of the track border can be interpreted as obstacles to avoid. The core idea in the perspective of F1Tenth is to find the largest gap in the LiDAR point cloud in front of the vehicle and set the target wheel angle towards it. Applying some minor modifications and integration of the non-holonomic constraint of a car-like vehicle and its dimensions, the method is able to control the RC car around a track in a robust way \cite{F1Tenth:2020a}.

\subsubsection{PID-like local tracking}

In the second group, the methods that operate with some kind of local path planning from the output of \textit{Perception} layer are classed. Since the preferred trajectory to follow is available, error terms can be calculated, and the controller that determines a corrective action can be applied. A common example of this method is the well-known PID controller.

\subsubsection{Optimization-based global tracking}

The last group of our summary contains the techniques where all of the layers are implemented. From the features of the \textit{Perception} layer, a map of the track is generated. Thus, a globally optimal trajectory can be calculated offline. Since, at the actual time frame, a part of the fixed global trajectory is extracted as local to follow, predictive control algorithms are worth utilizing. In autonomous driving term, one of the most relevant such methods is the model predictive control (MPC). This approach operates with a dynamic model of the system to predict its future states and optimize the inputs to minimize a given cost function, taking into account the fulfillment of the constraints, e.g., the limit of the track.

\section{Developed method for F1Tenth racing} \label{sec:our_methods}

\begin{figure}[!ht]
    \centering
    \includegraphics[width=0.40\textwidth]{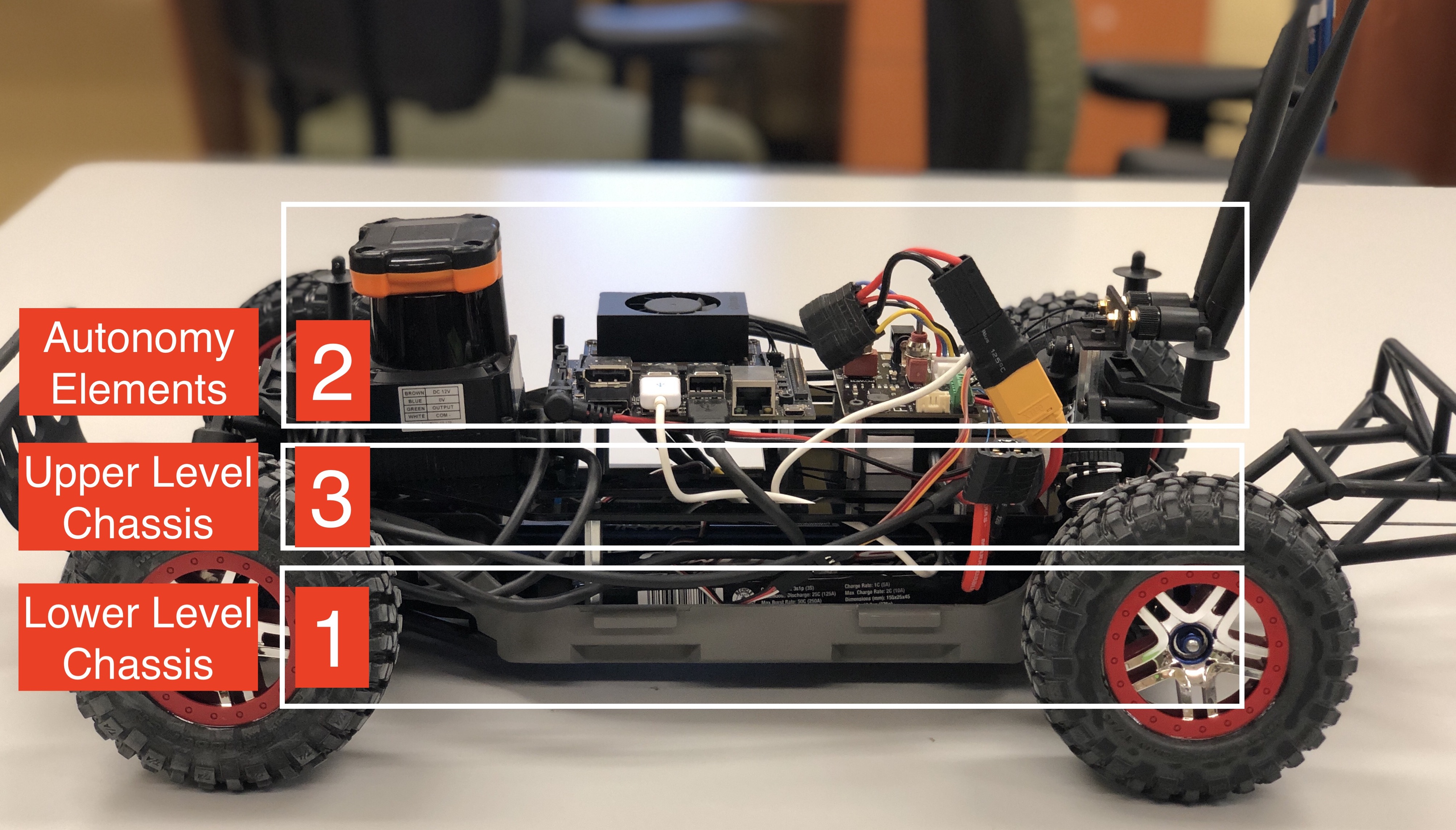}
    \caption{F1Tenth experimental vehicle \cite{F1Tenth:2020b}}
    \label{fig:f1tenth_vehicle}
\end{figure}

The examined control method belongs to the second PID-like local tracking group, thus there are two main parts: the path planning from the measured LiDAR point cloud and the determination of the target steering angle and vehicle velocity. Our implementation adapts to the specifications of F1Tenth competitions, where 1/10 scale electric motor-driven RC cars race on a track marked by pipes as walls. An example of a race can be found in Figure \ref{fig:f1tenth_vehicle}.

\subsection{Path planning}

The path planning algorithm is responsible for the determination of the centerline ahead of the vehicle. The distance from the pipes is measured with a LiDAR sensor thus the aim is to transform the raw point cloud into the represented centerline. The challenge of this path planning task is induced by the 2D type of LiDAR addition to the height of the walls because, in most of the bends, the end of one of the walls is invisible.

\subsubsection{Segment point cloud}

\begin{figure}[!ht]
    \centering
    \includegraphics[width=0.40\textwidth]{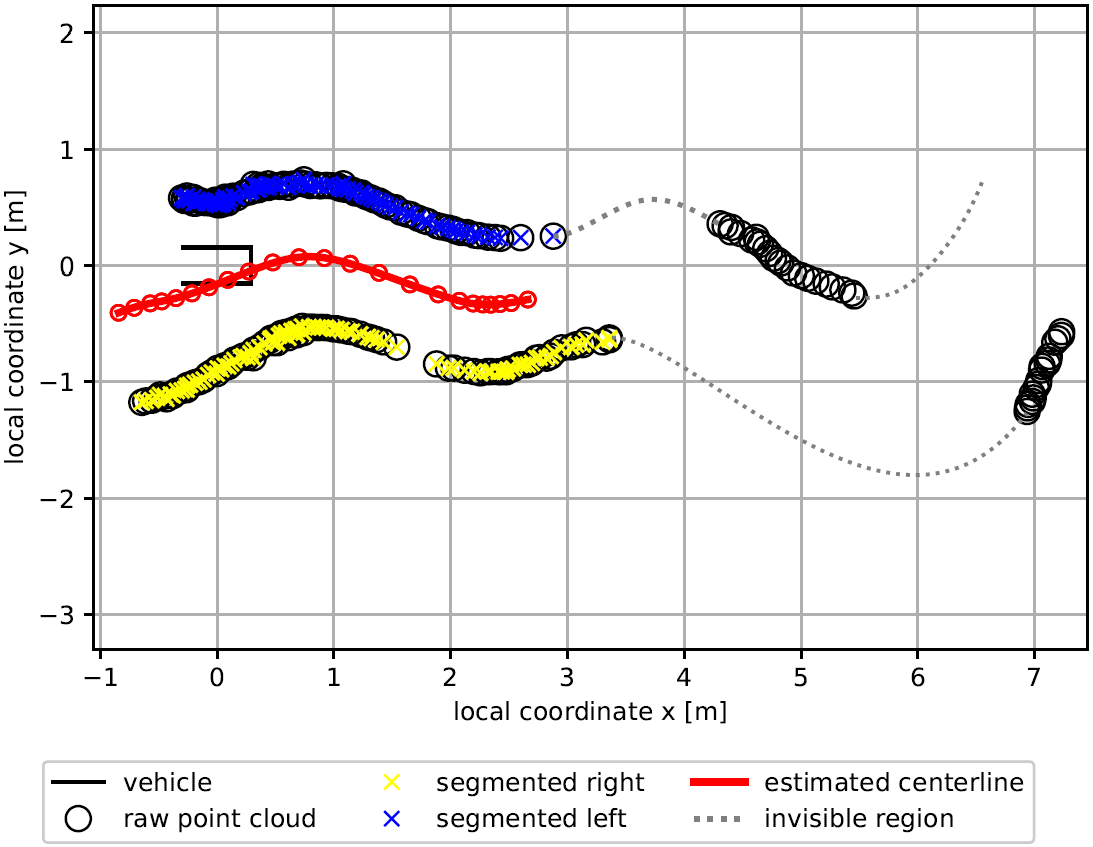}
    \caption{Segmentation of the raw point cloud measured by the LiDAR}
    \label{fig:path_planning_3}
\end{figure}

In the first part of the algorithm, a segmentation is performed as the two sides are only clearly separable from the raw point cloud in a straight route case. In most cases, several small parts of the walls are measured, or even only from one side (the points behind the vehicle are neglected). The separation is based on the weighted mean of the distance of the consecutive LiDAR measurements, an adaptive threshold is determined on previous measurements on numerous track layouts. An elimination algorithm is performed on the separated sections based on the length, distance from the vehicle, and angle of the sections, and the remains are handled as possible left and right walls. Finally, the segmented left/right sides are selected based on the count and pose of the remains. It is almost impossible for all of the possible sides to be eliminated, but in lots of cases, only one side is detected by the presented algorithm. An example of a slalom case can be found in Figure \ref{fig:path_planning_3}.

\subsubsection{Adaptive track width estimation}

\begin{figure}[!ht]
    \centering
    \includegraphics[width=0.40\textwidth]{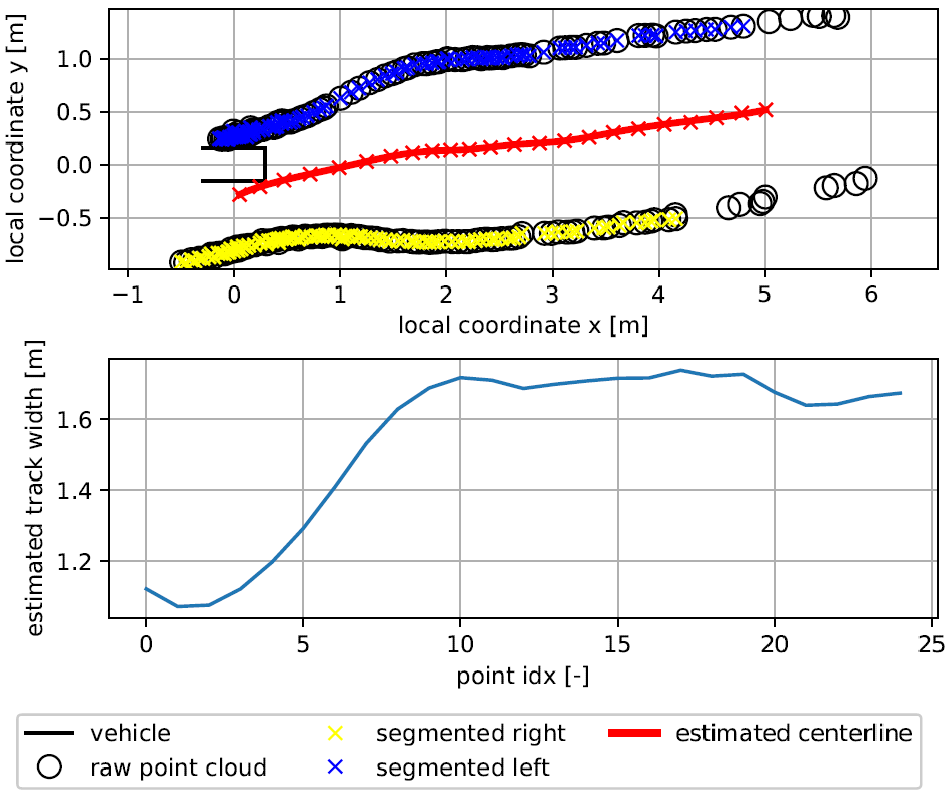}
    \caption{Track width estimation of the actual measurement}
    \label{fig:path_planning_2}
\end{figure}

A key parameter of our method is the actual track width. Since this can vary even inside an actual measurement, see upper plot of Figure \ref{fig:path_planning_2}, a point-by-point estimate is required. With a given resolution along the side, pairs are searched in a narrow-angle range. If the calculated distances are reliable, the track width vector is formulated with a moving window-based average. Figure \ref{fig:path_planning_2} demonstrates the process when the width changes rapidly, from 1 m to 2 m, but the estimator is able to follow.

\subsubsection{Generate centerline}

\begin{figure}[!ht]
    \centering
    \includegraphics[width=0.40\textwidth]{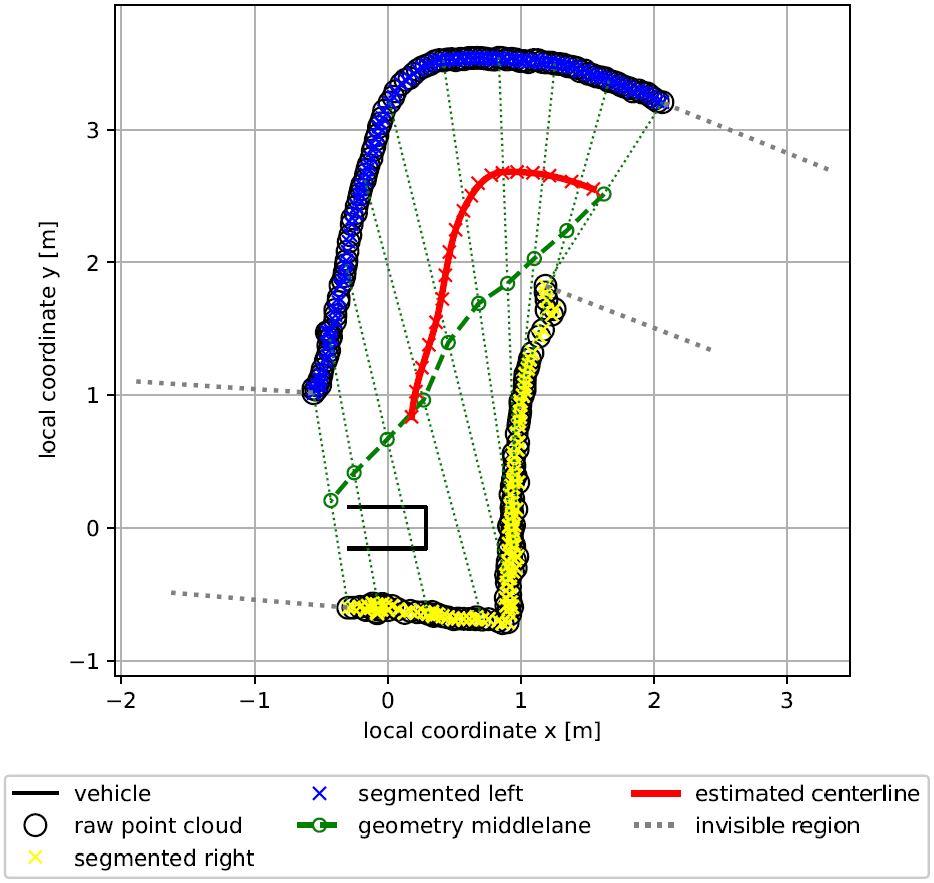}
    \caption{Example Track of an F1Tenth race}
    \label{fig:path_planning_1}
\end{figure}

Even if both sides can be detected, the geometric mean of the sides is not equal to the centerline of the track. An illustrative example can be found in Figure \ref{fig:path_planning_1}. Instead of the geometry mean, the more representative side is chosen and moved with the half-track width.
The illustrated case proves that this method is also appropriate when the geometry mean falls due to the invisible regions of the sides, and the track width can be estimated from just that small region, where both sides are visible.

This estimation is crucial for a local planner-based control algorithm because the chosen side can change even in consecutive timesteps, and a constant width value results in fluctuating centerline to follow, which induces an oscillation in the target control signals, resulting in instability.

\subsection{Improved Stanley-control for racing purposes}

\subsubsection{Lateral control}

For lateral control of the vehicle, a modified version of the Stanley controller \cite{hoffmann:2007} is integrated. The method operates by comparing the current pose of the vehicle with the calculated centerline trajectory to follow at the front axle of the car.

\begin{figure}[!ht]
    \centering
    \includegraphics[width=0.30\textwidth]{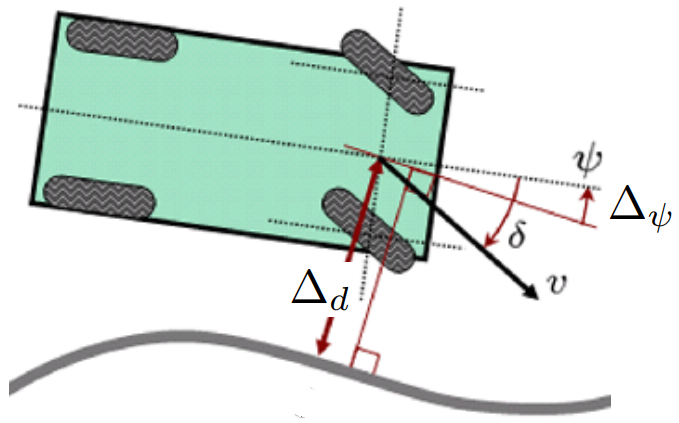}
    \caption{Illustration of the main error component of the Stanley-controller \cite{stanley_error}}
    \label{fig:stanley_error}
\end{figure}

The target wheel angle consists of two main components: the $\Delta_\psi$ angle error and $\Delta_d$ cross-track error, which are illustrated in Figure \ref{fig:stanley_error}. These errors are scaled with the proportional control components $k_{ang}$ and $k_{dist}$.  Applying the non-holonomic constraints of the vehicle in the examination of the decay of the lateral error, the value is divided with the $v$ vehicle velocity, and the $\arctan$ function is utilized. Furthermore, a $k_{soft}$ additive and $k_{damp}$ proportional components are also included in the denominator to guarantee the tracking performance both at low and high speeds.

\begin{subequations}
\begin{align}
    \Delta_\psi = {} &\psi(t) - \psi_{ss}(t) \\
    \begin{split}\label{}
    \Delta_d = {} & \cos{(\psi_{cp}(t))} \cdot (y_{cp}(t) - y_{ego}(t))\\
                  & - \sin{(\psi_{cp}(t))} \cdot (x_{cp}(t) - x_{ego}(t))
    \end{split}\\
    \Delta_r = {} & \dot\psi(t) - \dot\psi_{cp}(t)\\
    \Delta_\delta = {} & \delta_{measured,k}-\delta_{measured,k-1}
\end{align}
\end{subequations}

Using purely these two terms, only a kinematic relation is established between the orientation of the vehicle and the front wheels from the control perspective. However, controlling the dynamics can improve the tracking performance and could stabilize the movement as a damper. Thus, the yaw rate error is formulated as $\Delta_r$ and with a $k_{rate}$ gain factor is integrated into the control command. This is similar to the D component in the well-known PID control schema.

Finally, it is worth compensating the actuator delay and overshoot, which can be performed if $\Delta_\delta$ the time difference of the measured wheel angle is also scaled with $k_{steer}$.

The control input can be found in Equation \eqref{eq:control_input} summarizing the four: heading, cross-track, yaw rate, and steering components of the target wheel angle.
\begin{multline} \label{eq:control_input}
    \delta = \underbracket{k_{\text{ang}} \cdot \Delta_{\psi}}_\text{heading} + \underbracket{\arctan\Big({\frac{k_{\text{dist}} \cdot \Delta_d}{v \cdot k_{\text{damp}} + k_{\text{soft}}}}\Big)}_\text{cross track} + \\
    + \underbracket{k_{\text{rate}} \cdot \Delta_r}_\text{yaw rate} + \underbracket{k_{\text{steer}} \cdot \Delta_\delta}_\text{steering}
\end{multline}

\subsubsection{Longitudinal control}

The output of the path planning algorithm yields a sparse reference centerline, typically comprising approximately 50 points. To optimize this centerline for the control algorithm, a series of post-processing steps are employed. Initially, Laplacian smoothing is applied to the reference centerline to mitigate noise and irregularities. Subsequently, an Opheim simplification technique, integrated into the psimpl library \cite{ElmarKoning2013}, further refines the smoothed line. The resultant trajectory is then interpolated using a cubic $C^2$ spline \cite{TinoKluge2021} to ensure smoothness and continuity.

The core component of the longitudinal control algorithm is a minimum time velocity profile fitting method motivated by  \cite{Consolini2017}. The technique operates with a forward and backward iteration through the fitted spline and calculates the velocity of every $i$ trajectory point as,
\begin{subequations} \label{eq:longitudinal_1}
\begin{align}
    v_{g} &= \sqrt{{v_{i-1,f}}^2 + 2\cdot s_{i,f} \cdot a_{x,max}}, \\
    v_{i,f} &= \max(v_{min}, \min(v_{max}, v_{g}, v_{f,c}), \\
    v_{l} &= \sqrt{{v_{i+1,b}}^2 + 2\cdot s_{i,b} \cdot a_{x,min}}, \\
    v_{i,b} &= \max(v_{min}, \min(v_{max}, v_{l}, v_{c}), \\
    v_{i} &= \min(v_{i,f}, v_{i,b}),
\end{align}
\end{subequations}
where the $s$ values are geometry distances and the $v_{c}$ velocity is given by the lateral limit and the $\kappa_{i}$ curvature of the actual point,
\begin{subequations} \label{eq:longitudinal_2}
\begin{align}
    s_{i,f} &= \sqrt{(p_{i,x}-p_{i-1,x})^2+(p_{i,y}-p_{i-1,y})^2}, \\
    s_{i,b} &= \sqrt{(p_{i+1,x}-p_{i,x})^2+(p_{i+1,y}-p_{i,y})^2}, \\
    v_{c} &= \sqrt{a_{y,max} / |\kappa_{i}|}.
\end{align}
\end{subequations}
Finally, the continuity of the vehicle velocity and the stable motion are guaranteed with the $\Delta_{a_{min}}$ and $\Delta_{a_{max}}$ allowed velocity change limits thus the target control velocity is,
\begin{equation}
    v_{k+1} = \max(v_{k,meas} + \Delta_{a,min},\min(v_{k,meas} + \Delta_{a,max}, v_i)).
\end{equation}
In Equations \eqref{eq:longitudinal_1} and \eqref{eq:longitudinal_2}, the $v_{min}$, $v_{max}$ and $a_{x,min}$, $a_{x,max}$, $a_{y,max}$ are the minimum and maximum longitudinal and lateral velocity and acceleration parameters, respectively.

\subsubsection{Adaptive lookahead point}

Finally, the original method is improved considering the racing application. Because in this context, minimizing lap time is paramount, we do not want to correct exactly the actual deviations, instead take the corners as sharp as possible.

This induces that a point ahead of the vehicle should be tracked, therefore the control point at the front axle is transformed into an adaptive lookahead point. The determination of the actual lookahead distance relies on the curvature of the trajectory and configurable parameters defining the $L_{max}$ maximum value and curvature normalization factor. These parameters should be adapted to the specific characteristics of the track layout. Based on the calculated lookahead distance, a point is dynamically selected along the trajectory to serve as the lookahead point.

\section{F1Tenth experimental vehicle} \label{sec:vehicle}

The method is developed for the F1TENTH autonomous racing platform \cite{Kelly:2020}. This is built around the 1/10 size electric motor-driven Traxxas Slash 4x4 Ultimate \cite{Traxxas}. The vehicle is all-wheel-driven, and the electric motor is connected to the driveshaft at the center of the vehicle, which transfers the motion to the differential gears in the axes. The wheels are connected to the chassis with double-wishbone suspensions. The steering mechanism is a basic Ackermann-steering geometry.

The autonomous stack consists of a Hokuyo UTM-30LX LiDAR as the main environment sensor for measurement of the pipes around the track, from vehicle sensor a Bosch BMI160 MEMS IMU, and an encoder to measure the motor speed. The steering actuator is a Traxxas High-Torque 600 Brushless Digital Servo. The developed algorithms run on a NVIDIA Jetson AGX Orin computer with ROS2 middleware framework in C++ and Python programming language.

\section{Results} \label{sec:results}

The results of the proposed control method are presented in two parts: first, a brief summary is illustrated of a tuning process to minimize lap time on an F1Tenth race, and subsequently, the result is evaluated compared to other control techniques outlined in Section \ref{control_methods}.

\subsection{Tuning process of the proposed method}

The paramount aim in this F1Tenth autonomous racing is to finish the lap as fast as possible without crashing into the walls. For the proposed control method, this can be achieved in two ways. One is to increase the $v$ velocity limit parameters as high as possible and ensure trajectory tracking with the continuous tuning of the steering control parameters, and the other is to shorten the distance traveled by cutting the bends if possible. 
\begin{figure}[!ht]
    \centering
    \includegraphics[width=0.40\textwidth]{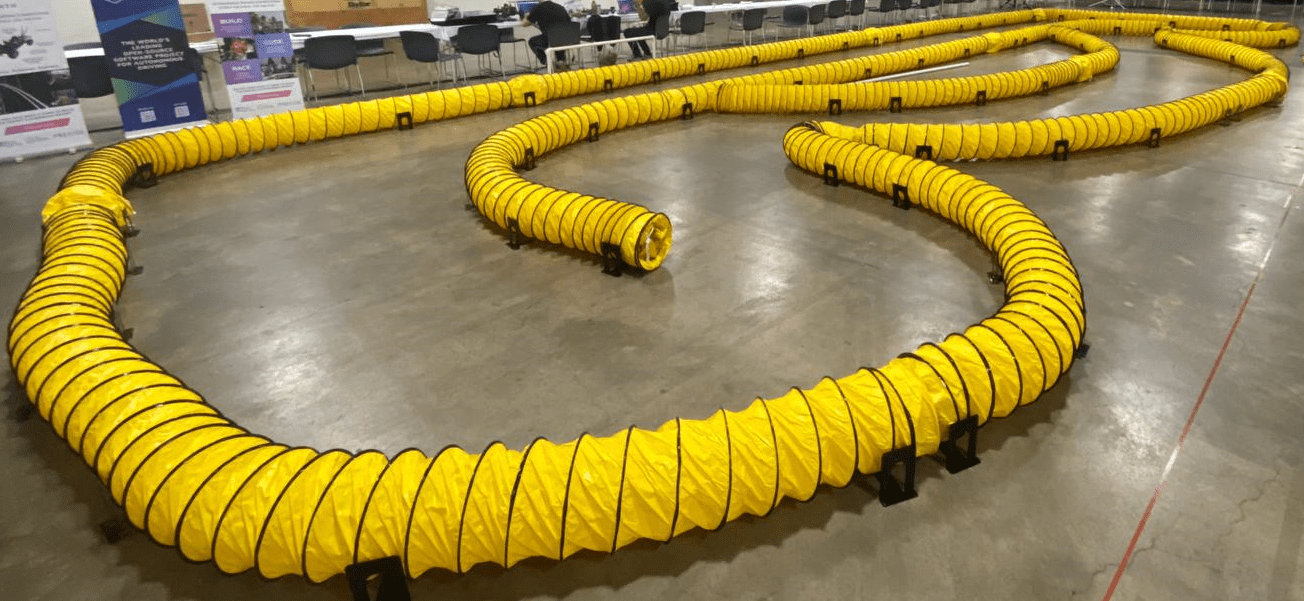}
    \caption{F1Tenth example track layout}
    \label{fig:track_layout}
\end{figure}
The layout of the track can be found in Figure \ref{fig:track_layout}, the average width is 7 times wider than the vehicle, thus, it is clear that the centerline is far from the optimal path.

A general robust setting, as a base setup, has been determined for the control parameters with the following values,
\begin{subequations}
\begin{equation}
    k_{ang} = 0.6, \,\ k_{dist}=0.5, \,\ k_{soft}=5.0, \,\ k_{damp}=1.0,
\end{equation}
\begin{equation}
    k_{yaw}=-0.013, \,\ v_{min} = 2.0, \,\ v_{max}=4.0, \,\ L_{max}=0.2.
\end{equation}
\end{subequations}
With this setting, the lap time is $17.5 \ s$, and the vehicle almost perfectly follows the centerline.

\begin{figure}[!ht]
    \centering
    \includegraphics[width=0.45\textwidth]{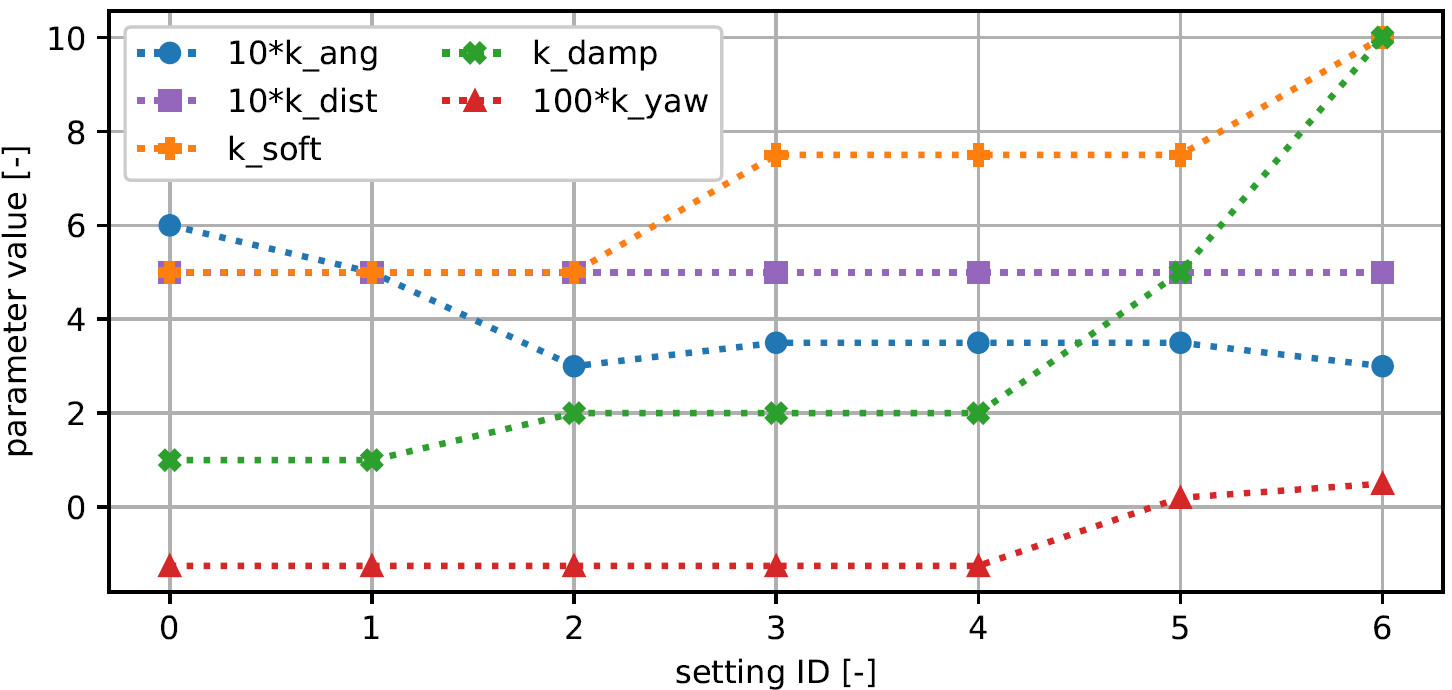}
    \caption{Steering control parameters of different setting cases}
    \label{fig:result_control}
\end{figure}

To achieve the optimum setting, many tuning steps have been carried out, of which a brief summary with only 6 settings is presented. In the first two steps, the $L_{max}$ lookahead distance is increased to around $0.6 \ m$ to induce traveled distance reduction with sharper cornering. In parallel, the $v_{max}$ maximum allowed velocity limit is increased from $4 \ m/s$ to $6.5 \ ms/s$, and also a bit higher $v_{min}$ minimum speed is used as $2.3 \ m/s$. The lap time decreases with $11\%$, it is $1.9 \ s$ lower than with the base setting. Since the control point is moved forward with a vehicle length in these cases, the angle error becomes significantly higher. To eliminate the crash into the inner corner in the bends, the $k_{ang}$ gain value is decreased to $0.3$ as illustrated in Figure \ref{fig:result_control}.

\begin{figure}[!ht]
    \centering
    \includegraphics[width=0.45\textwidth]{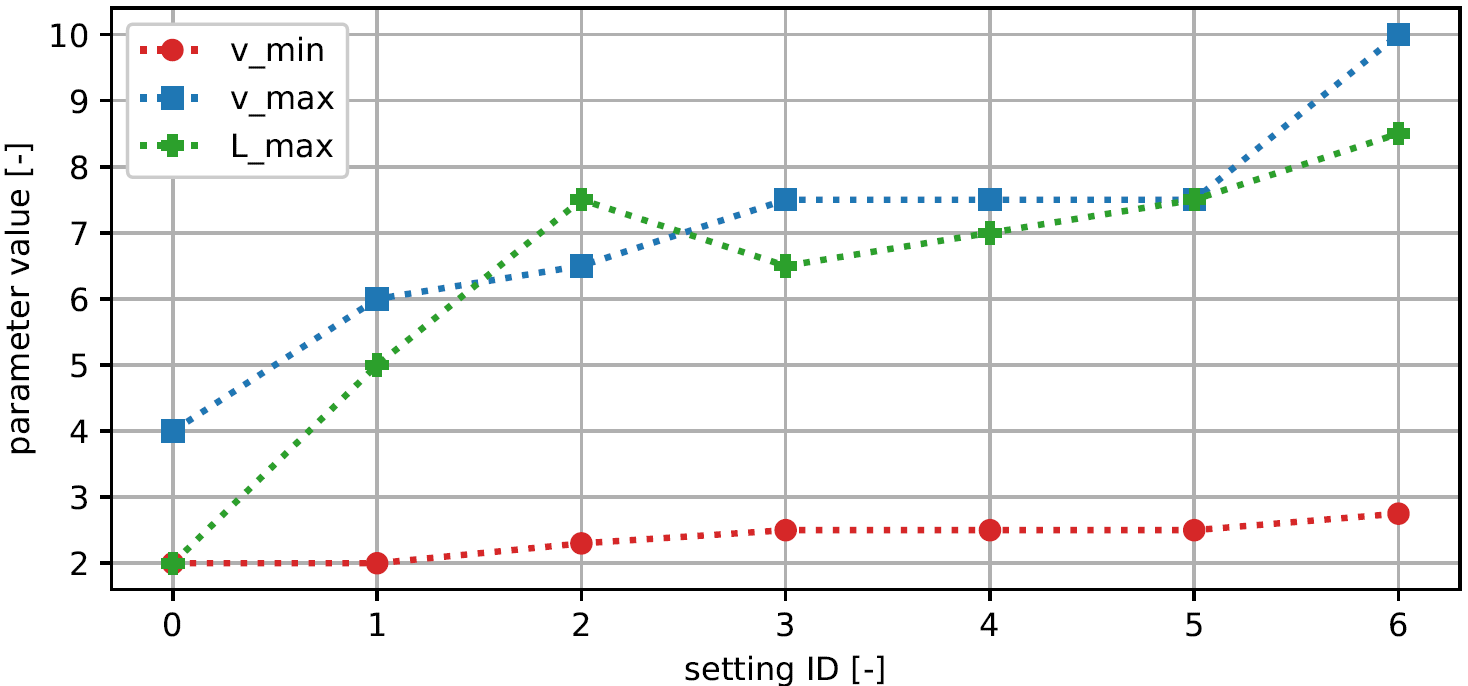}
    \caption{Motion control parameters of different setting cases}
    \label{fig:result_speed}
\end{figure}

Thereafter, in setting 3 and 4, the aim is to stabilize driving as at $6 \ m/s$, the high-speed range is beginning at this 1/10 scaled vehicle. The stability can be emphasized with the increase of the $k_{soft}$ and $k_{damp}$ factors. Although it is illustrated in Figure \ref{fig:result_speed} that the velocity limits are the same, the lap time decreases with another $1.3 \ s$ to $14.3 \ s$ because the oscillation is eliminated.

The final parameter to tune is the $k_{rate}$ component. Until now, the yaw rate error is integrated as a negative feedback as daming. However, in the final case, the $k_{rate}$ is changed to a low positive value which prefers more aggressive cornering. This allows both higher speed limits and increased lookahead distance, while the stability is guaranteed by further tuning of the $k_{soft}$,$k_{damp}$ parameters. The optimal parameter setting is,
\begin{subequations}
\begin{equation}
    k_{ang} = 0.30, \,\ k_{dist}=0.5, \,\ k_{soft}=10.0, \,\ k_{damp}=10.0,
\end{equation}
\begin{equation}
    k_{yaw}= 0.005, \,\ v_{min} = 2.75, \,\ v_{max}=10.0, \,\ L_{max}=0.85.
\end{equation}
\end{subequations}
The interesting fact is that in all cases, $k_{dist}$ remains at the base value, which can be explained by the lesser importance of the cross-track error in this racing application.

\begin{figure}[!ht]
    \centering
    \includegraphics[width=0.45\textwidth]{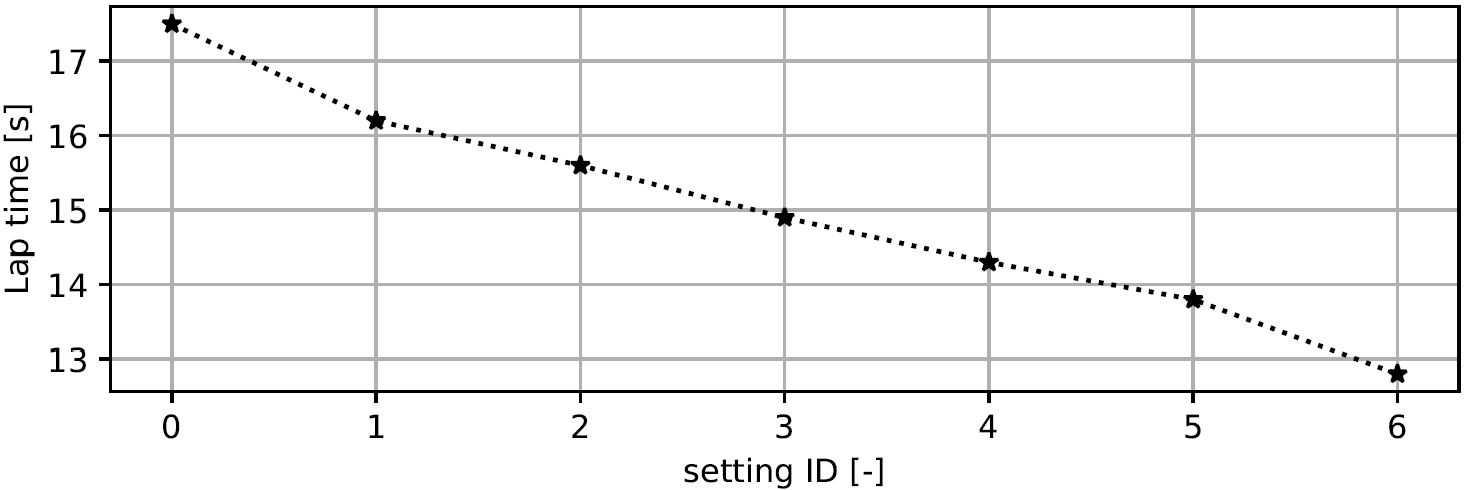}
    \caption{Lap time of the different setting cases}
    \label{fig:result_time}
\end{figure}

It should be emphasized that the control parameters are overturned to the actual track, but as it is demonstrated in Figure \ref{fig:result_time}, the lap time is significantly reduced to $12.8 \ s$ that is $28\%$ decrease in relative term from the base value.
The lap time signal throughout the tuning process could not be as smooth as it is illustrated, but only the main changes that are stable and robust are examined.

\subsection{Comparison with other methods on a real race}

In this paper, the evaluation of the proposed controller method is also a main goal, which requires a comparison with other techniques. The proposed controller was used in the F1Tenth competition of the IROS23 conference, where all of the mentioned methods in Section \ref{control_methods} were represented.

From personal discussion, it was clear that both reactive and optimisation-based control methods were applied. The teams with the former could reach around $16 \ s$ lap time, while the latter was implemented by the winning team, whose best time was $11.82 \ s$. This winner time is only $8\%$ lower than the best of our, futhermore with the proposed method $12.4 \ s$ could be reached as well but it was no longer robust.

Assuming that everyone has tuned their algorithm to the actual track, two conclusions can be stated. The first is a PID-like local tracking method that has significantly higher performance than a reactive one despite the fact that both operate only with the actual LiDAR measurement without any map. Second, although the $8\%$ in lap time is an enormous difference in real-sized car racing, but from a control perspective, it demonstrates that a properly tuned PID-like method without any map or optimization can be comparable with a state-of-the-art method that operates with a map, global racing line, and complex optimization in the computation of control target signals.

\section{Conclusion} \label{sec:conclusion}

In this paper, the development of a local path planning-based autonomous control algorithm and its evaluation in RC car racing were presented. The advantage of the proposed method is that it does not require any pre-calculated complex mapping phase.

The results in Figure \ref{fig:result_time} illustrate the capability of the proposed method since the base setting, which can guarantee robust trajectory tracking in any kind of track, can be tuned further to the actual track, resulting in $28\%$ lap time reduction. In the context of control techniques, this performance level is only $8\%$ worse than the state-of-the-art method, which requires map generation before. Nevertheless, the outcomes of the proposed technique can have a higher impact on the automotive industry because of the closer similarity to the real-life scenarios.

\bibliographystyle{IEEEtran}
\bibliography{references}

\end{document}